# Nonmonotonic Probabilistic Logics between Model-Theoretic Probabilistic Logic and Probabilistic Logic under Coherence

Thomas Lukasiewicz*
Dipartimento di Informatica e Sistemistica,
Università di Roma "La Sapienza"
Via Salaria 113, 00198 Roma, Italy
lukasiewicz@dis.uniroma1.it

## Abstract

Recently, it has been shown that probabilistic entailment under coherence is weaker than model-theoretic probabilistic entailment. Moreover, probabilistic entailment under coherence is a generalization of default entailment in System $P$. In this paper, we continue this line of research by presenting probabilistic generalizations of more sophisticated notions of classical default entailment that lie between model-theoretic probabilistic entailment and probabilistic entailment under coherence. That is, the new formalisms properly generalize their counterparts in classical default reasoning, they are weaker than model-theoretic probabilistic entailment, and they are stronger than probabilistic entailment under coherence. The new formalisms are useful especially for handling probabilistic inconsistencies related to conditioning on zero events. They can also be applied for probabilistic belief revision. More generally, in the same spirit as a similar previous paper, this paper sheds light on exciting new formalisms for probabilistic reasoning beyond the well-known standard ones.

## 1 Introduction

In the recent decades, probabilistic reasoning has become an important research topic in artificial intelligence. In particular, extensive research has been carried out on probabilistic reasoning with interval restrictions for conditional probabilities, also called *conditional constraints* [40].

One important approach for handling conditional constraints is model-theoretic probabilistic logic, which comes especially from philosophy and logic, and whose roots go back to Boole's book of 1854 "The Laws of Thought" [11]. There is a wide spectrum of formal languages that have been explored in model-theoretic probabilistic logic, ranging from constraints for unconditional and conditional events [16, 20, 2, 53, 19, 23, 33, 45, 39, 40, 42, 46] to rich languages that specify linear inequalities over events [22]. The main problems related to model-theoretic probabilistic logic are checking satisfiability, deciding logical consequence, and computing tight logically entailed intervals.

Another important approach to probabilistic reasoning with conditional constraints, which has been extensively explored in the area of statistics, is based on the coherence principle of de Finetti and suitable generalizations of it [8, 12, 13, 14, 15, 27, 28, 29, 50], or on similar principles that have been adopted for lower and upper probabilities [49, 54]. Two important aspects of dealing with uncertainty in this framework are (i) checking the consistency of a probabilistic assessment, and (ii) the propagation of a given assessment to further uncertain quantities.

Recently, the relationship between model-theoretic probabilistic logic and probabilistic logic under coherence has been thoroughly explored in [10]. In particular, it turned out that probabilistic entailment under coherence is weaker than model-theoretic probabilistic entailment. Moreover, it has been shown in [10] that probabilistic entailment under coherence is a generalization of classical default entailment in System $P$, while model-theoretic probabilistic entailment is well-known to be a generalization of model-theoretic entailment in classical propositional logics.

In this paper, we continue this line of research by presenting probabilistic generalizations of more sophisticated formalisms for default reasoning from conditional knowledge bases, which lie between probabilistic entailment under coherence and model-theoretic probabilistic entailment. That is, the new notions of entailment generalize their classical counterparts, they are stronger than entailment under coherence, and weaker than model-theoretic entailment.

---

The literature contains several different proposals for default reasoning from conditional knowledge bases and extensive work on its desired properties. The core of these properties are the rationality postulates of System $P$ proposed by Kraus et al. [34]. It turned out that these rationality postulates constitute a sound and complete axiom system for several classical model-theoretic entailment relations under uncertainty measures on worlds. In detail, they characterize classical model-theoretic entailment under preferential structures [51, 34], infinitesimal probabilities [1, 47], possibility measures [17], and world rankings [52, 31]. They also characterize an entailment relation based on conditional objects [18]. A survey of all these relationships is given in [6, 24].

Mainly to solve problems with irrelevant information, the notion of rational closure as a more adventurous notion of entailment was introduced by Lehmann [36, 38]. It is equivalent to entailment in System $Z$ by Pearl [48], to the least specific possibility entailment by Benferhat et al. [5], and to a conditional (modal) logic-based entailment by Lamarre [35]. Finally, mainly to solve problems with property inheritance from classes to exceptional subclasses, the maximum entropy approach to default entailment was proposed by Goldszmidt et al. [30]; lexicographic entailment was introduced by Lehmann [37] and Benferhat et al. [4]; conditional entailment was proposed by Geffner [25, 26]; and an infinitesimal belief function approach was suggested by Benferhat et al. [7].

In this paper, we define probabilistic generalizations of entailment in Pearl's System $Z$ and Lehmann's lexicographic entailment that lie between entailment in model-theoretic probabilistic logic and entailment in probabilistic logic under coherence. Roughly speaking, the main difference between model-theoretic probabilistic entailment and probabilistic entailment under coherence is that the former realizes an inheritance of logical knowledge, while the latter does not. Intuitively, the new formalisms now add a strategy for resolving inconsistencies to model-theoretic probabilistic logic, and a restricted form of inheritance of logical knowledge to probabilistic logic under coherence. This is why the new notions of entailment are weaker than entailment in model-theoretic probabilistic logic and stronger than entailment in probabilistic logic under coherence.

Hence, the new formalisms are refinements of both model-theoretic probabilistic entailment and probabilistic entailment under coherence. They can be used in place of model-theoretic probabilistic entailment, when we want to resolve probabilistic inconsistencies related to *conditioning on zero events*. Here, the new formalisms are especially well-suited as they coincide with model-theoretic probabilistic entailment as long as we condition on non-zero events. Furthermore, the new formalisms can be used in place of probabilistic entailment under coherence, when we also want to have an inheritance of logical knowledge. Finally, another important application is probabilistic belief revision.

A companion paper [44] presents similar probabilistic generalizations of entailment in Pearl's System $Z$ and of Lehmann's lexicographic entailment. These formalisms, however, are quite different from the ones in this paper. In detail, the new formalisms in [44] add an inheritance of probabilistic knowledge and a strategy for resolving inconsistencies to entailment in model-theoretic probabilistic logic. For this reason, they are generally much stronger than entailment in model-theoretic probabilistic logic. Thus, they are especially useful where the classical notion of model-theoretic entailment is too weak, for example, in probabilistic logic programming [42, 41]. Other applications are deriving degrees of belief from statistical knowledge and degrees of belief, handling inconsistencies in probabilistic knowledge bases, and probabilistic belief revision.

The main contributions of this paper are as follows:

• We show that the notions of g-coherence and of g-coherent entailment in probabilistic logic under coherence can be expressed in terms of probability rankings.

• We present new probabilistic generalizations of entailment in Pearl's System $Z$ and Lehmann's lexicographic entailment, which lie between model-theoretic probabilistic entailment and probabilistic entailment under coherence.

• We analyze and compare some general nonmonotonic properties of the new notions of entailment, of model-theoretic probabilistic entailment, and of probabilistic entailment under coherence. In particular, it turns out that the new formalisms satisfy the properties of Rational Monotonicity and Irrelevance, while probabilistic entailment under coherence is lacking these properties.

• We analyze the relationship between the new notion of entailment in System $Z$, the new notion of lexicographic entailment, model-theoretic probabilistic entailment, and probabilistic entailment under coherence.

• We show that the new notions of entailment in System $Z$ and of lexicographic entailment are proper generalizations of their classical counterparts.

The rest of this paper is organized as follows. Section 2 gives some technical preliminaries. In Section 3, we show that probabilistic reasoning under coherence can be expressed in terms of probability rankings, and we introduce the new probabilistic notions of entailment in System $Z$ and of lexicographic entailment. Section 4 concentrates on the properties of the new formalisms and of the standard ones. In Section 5, we summarize the main results and give an outlook on future research. Note that detailed proofs of all results are given in the extended paper [43].

## 2 Preliminaries

In this section, we first define probabilistic knowledge bases. We then recall the notions of satisfiability and of logical entailment from model-theoretic probabilistic logic, and the notions of g-coherence and of g-coherent entailment from probabilistic logic under coherence.

### 2.1 Probabilistic Knowledge Bases

We assume a set of *basic events* $\Phi = \{p_1, \ldots, p_n\}$ with $n \geq 1$. We use $\bot$ and $\top$ to denote *false* and *true*, respectively. We define *events* by induction as follows. Every element of $\Phi \cup \{\bot, \top\}$ is an event. If $\phi$ and $\psi$ are events, then also $\neg \phi$ and $(\phi \wedge \psi)$. A *logical constraint* is an event of the form $\psi \Leftarrow \phi$. A *conditional event* is an expression of the form $\psi | \phi$ with events $\psi$ and $\phi$. A *conditional constraint* is an expression of the form $(\psi | \phi)[l, u]$ with events $\psi, \phi$, and real numbers $l, u \in [0, 1]$. We define *probabilistic formulas* by induction as follows. Every conditional constraint is a probabilistic formula. If $F$ and $G$ are probabilistic formulas, then also $\neg F$ and $(F \wedge G)$. We use $(F \vee G)$ and $(F \Leftarrow G)$ to abbreviate $\neg(\neg F \wedge \neg G)$ and $\neg(\neg F \wedge G)$, respectively, where $F$ and $G$ are either two events or two probabilistic formulas. We adopt the usual conventions to eliminate parentheses.

A *world* $I$ is a truth assignment to the basic events in $\Phi$ (that is, a mapping $I \colon \Phi \to \{\textbf{true}, \textbf{false}\}$), which is inductively extended to all events by $I(\bot) = \textbf{false}$, $I(\top) = \textbf{true}$, $I(\neg \phi) = \textbf{true}$ iff $I(\phi) = \textbf{false}$, and $I((\phi \wedge \psi)) = \textbf{true}$ iff $I(\phi) = I(\psi) = \textbf{true}$. We use $\mathcal{I}_\Phi$ to denote the set of all worlds for $\Phi$. A world $I$ *satisfies* an event $\phi$, or $I$ is a *model* of $\phi$, denoted $I \models \phi$, iff $I(\phi) = \textbf{true}$. We extend worlds $I$ to conditional events $\psi | \phi$ by $I(\psi | \phi) = \textbf{true}$ iff $I \models \psi \wedge \phi$, $I(\psi | \phi) = \textbf{false}$ iff $I \models \neg \psi \wedge \phi$, and $I(\psi | \phi) = \textbf{indeterminate}$ iff $I \models \neg \phi$. A *probabilistic interpretation* $Pr$ is a probability function on $\mathcal{I}_\Phi$ (that is, a mapping $Pr \colon \mathcal{I}_\Phi \to [0, 1]$ such that all $Pr(I)$ with $I \in \mathcal{I}_\Phi$ sum up to 1). The *probability* of an event $\phi$ in the probabilistic interpretation $Pr$, denoted $Pr(\phi)$, is the sum of all $Pr(I)$ such that $I \in \mathcal{I}_\Phi$ and $I \models \phi$. For events $\phi$ and $\psi$ with $Pr(\phi) > 0$, we write $Pr(\psi | \phi)$ to abbreviate $Pr(\psi \wedge \phi) / Pr(\phi)$. The *truth* of logical constraints and probabilistic formulas $F$ in a probabilistic interpretation $Pr$, denoted $Pr \models F$, is defined as follows:

- $Pr \models \psi \Leftarrow \phi$ iff $Pr(\psi \wedge \phi) = Pr(\phi)$.
- $Pr \models (\psi | \phi)[l, u]$ iff $Pr(\phi) = 0$ or $Pr(\psi | \phi) \in [l, u]$.
- $Pr \models \neg F$ iff not $Pr \models F$.

- $Pr \models (F \wedge G)$ iff $Pr \models F$ and $Pr \models G$.

We say $Pr$ *satisfies* $F$, or $Pr$ is a *model* of $F$, iff $Pr \models F$. We say $Pr$ *satisfies* a set of logical and conditional constraints $\mathcal{F}$, or $Pr$ is a *model* of $\mathcal{F}$, denoted $Pr \models \mathcal{F}$, iff $Pr$ is a model of all $F \in \mathcal{F}$.

A *probabilistic knowledge base* $KB = (L, P)$ consists of a finite set of logical constraints $L$ and a finite set of conditional constraints $P$ such that (i) $l \leq u$ for all $(\psi | \phi)[l, u] \in P$, and (ii) $\psi_1 | \phi_1 \neq \psi_2 | \phi_2$ for any two distinct $(\psi_1 | \phi_1)[l_1, u_1], (\psi_2 | \phi_2)[l_2, u_2] \in P$. Note that here $\psi_1 | \phi_1 \neq \psi_2 | \phi_2$ iff either $\psi_1 \wedge \phi_1$ is not logically equivalent to $\psi_2 \wedge \phi_2$, or $\phi_1$ is not logically equivalent to $\phi_2$.

**Example 2.1** A sample probabilistic knowledge base $KB = (L, P)$ is given as follows:

$$L = \{\textsf{bird} \Leftarrow \textsf{penguin}\},$$
$$P = \{(\textsf{have\_legs} \,|\, \textsf{bird})[1, 1], (\textsf{fly} \,|\, \textsf{bird})[1, 1]\}.$$

Here, $L$ represents the logical knowledge "all penguins are birds". Moreover, in model-theoretic probabilistic logic, $P$ expresses the *logical knowledge* "all birds have legs" and "all birds fly", while in probabilistic logic under coherence and in the new nonmonotonic logics, $P$ expresses the *default logical knowledge* "typically, birds have legs" and "typically, birds fly".

Another probabilistic knowledge base $KB' = (L', P')$ is given as follows:

$$L' = \{\textsf{bird} \Leftarrow \textsf{penguin}\},$$
$$P' = \{(\textsf{have\_legs} \,|\, \textsf{bird})[1, 1], (\textsf{fly} \,|\, \textsf{bird})[1, 1],$$
$$(\textsf{fly} \,|\, \textsf{penguin})[0, 0.05]\}.$$

In model-theoretic probabilistic logic, the additional conditional constraint $(\textsf{fly} \,|\, \textsf{penguin})[0, 0.05]$ represents the *probabilistic knowledge* "at most 5% of all penguins fly", while in probabilistic logic under coherence and in the new nonmonotonic logics, $(\textsf{fly} \,|\, \textsf{penguin})[0, 0.05]$ represents the *default probabilistic knowledge* "typically, penguins fly with a probability of at most 0.05". $\Box$

### 2.2 Model-Theoretic Probabilistic Logic

We now recall the model-theoretic notions of satisfiability and of logical entailment.

A set of logical and conditional constraints $\mathcal{F}$ is *satisfiable* iff a model of $\mathcal{F}$ exists. A probabilistic knowledge base $KB = (L, P)$ is *satisfiable* iff $L \cup P$ is satisfiable.

A conditional constraint $C$ is a *logical consequence* of a set of logical and conditional constraints $\mathcal{F}$, denoted $\mathcal{F} \models C$, iff every model of $\mathcal{F}$ is also a model of $C$. A conditional constraint $(\psi|\phi)[l, u]$ is a *tight logical consequence* of $\mathcal{F}$, denoted $\mathcal{F}\models_{tight}(\psi|\phi)[l, u]$, iff $l$ (resp., $u$) is the infimum (resp., supremum) of $Pr(\psi|\phi)$ subject to all models $Pr$ of $\mathcal{F}$ with $Pr(\phi) > 0$. Note that we define $l = 1$ and $u = 0$, when $\mathcal{F}\models(\phi|\top)[0, 0]$. A conditional constraint $C$ is a *logical consequence* of a probabilistic knowledge base $KB$, denoted $KB \models C$, iff $L \cup P \models C$. It is a *tight logical consequence* of $KB$, denoted $KB\models_{tight}C$, iff $L \cup P\models_{tight}C$.

**Example 2.2** The probabilistic knowledge base $KB$ of Example 2.1 has the following logical consequences:

(fly | bird)[1, 1], (have_legs | bird)[1, 1],
(fly | penguin)[1, 1], (have_legs | penguin)[1, 1].

In fact, all these conditional constraints are even tight logical consequences of $KB$.

The probabilistic knowledge base $KB' = (L', P')$ of Example 2.1 has the following tight logical consequences:

(fly | bird)[1, 1], (have_legs | bird)[1, 1],
(fly | penguin)[1, 0], (have_legs | penguin)[1, 0].

Here, the empty interval "[1, 0]" in the last two conditional constraints is due to the fact that the logical property of being able to fly is inherited from birds to penguins and is incompatible there with the knowledge that at most 5% of all penguins are able to fly. Hence, in this example, the notion of logical entailment is too strong, as the desirable tight conclusions from $KB'$ are given by (fly | penguin)[0, 0.05] and (have_legs | penguin)[1, 1], respectively. □

### 2.3 Probabilistic Logic under Coherence

We now recall the notions of g-coherence and of g-coherent entailment. We define them by using some recent characterizations through concepts from default reasoning [10]. We first give some preparative definitions.

A probabilistic interpretation $Pr$ *verifies* a conditional constraint $(\psi|\phi)[l, u]$ iff $Pr(\phi) > 0$ and $Pr \models (\psi|\phi)[l, u]$. It *falsifies* a conditional constraint $(\psi|\phi)[l, u]$ iff $Pr(\phi) > 0$ and $Pr \not\models (\psi|\phi)[l, u]$. A set of conditional constraints $P$ *tolerates* a conditional constraint $C$ *under* a set of logical constraints $L$ iff $L \cup P$ has a model that verifies $C$. A set of conditional constraints $P$ is *under $L$ in conflict* with $C$ iff no model of $L \cup P$ verifies $C$.

A *conditional constraint ranking* $\sigma$ on a probabilistic knowledge base $KB = (L, P)$ maps each element of $P$ to a nonnegative integer. It is *admissible* with $KB$ iff every $P' \subseteq P$ that is under $L$ in conflict with some $C \in P$ contains a conditional constraint $C'$ such that $\sigma(C') < \sigma(C)$.

We are now ready to define the concept of g-coherence. A probabilistic knowledge base $KB$ is *g-coherent* iff there exists a conditional constraint ranking on $KB$ that is admissible with $KB$.

We next define the notion of g-coherent entailment. Let $KB = (L, P)$ be a g-coherent probabilistic knowledge base, and let $(\psi|\phi)[l, u]$ be a conditional constraint. Then, $(\psi|\phi)[l, u]$ is a *g-coherent consequence* of $KB$, denoted $KB \mathrel{\|\!\sim}^g (\psi|\phi)[l, u]$, iff $(L, P \cup \{(\psi|\phi)[p, p]\})$ is not g-coherent for all $p \in [0, l) \cup (u, 1]$. We say $(\psi|\phi)[l, u]$ is a *tight g-coherent consequence* of $KB$, denoted $KB \mathrel{\|\!\sim}^g_{tight} (\psi|\phi)[l, u]$, iff $l$ (resp., $u$) is the infimum (resp., supremum) of $p$ subject to all g-coherent probabilistic knowledge bases $(L, P \cup \{(\psi|\phi)[p, p]\})$.

**Example 2.3** The probabilistic knowledge base $KB$ of Example 2.1 has the following tight g-coherent consequences:

(fly | bird)[1, 1], (have_legs | bird)[1, 1],
(fly | penguin)[0, 1], (have_legs | penguin)[0, 1].

Here, the interval "[0, 1]" in the last two conditional constraints is due to the fact that the logical properties of being able to fly and of having legs are not inherited from birds to penguins. Thus, in this example, the notion of g-coherent entailment is too weak, as the desirable tight conclusions from $KB$ are given by (fly | penguin)[1, 1] and (have_legs | penguin)[1, 1], respectively.

The probabilistic knowledge base $KB'$ of Example 2.1 has the following tight g-coherent consequences:

(fly | bird)[1, 1], (have_legs | bird)[1, 1],
(fly | penguin)[0, 0.05], (have_legs | penguin)[0, 1].

Here, the last interval "[0, 1]" is due to the fact that the logical property of having legs is not inherited from birds to penguins. Thus, also in this example, g-coherent entailment is too weak, as the desirable tight conclusion from $KB'$ is given by (have_legs | penguin)[1, 1]. □

## 3 Nonmonotonic Probabilistic Logics

In this section, we first show that the notions of g-coherence and of g-coherent entailment can be expressed in terms of probability rankings. We then introduce new probabilistic generalizations of Pearl's entailment in System $Z$ and of Lehmann's lexicographic entailment. Observe that the notion of g-coherent entailment is defined through a set of probability rankings, while the new notions of entailment in System $Z$ and of lexicographic entailment are each defined through a unique single probability ranking.

In some sense, the new formalisms are second-order formalisms, as probability rankings are closely related to possibility measures over probabilities and can thus be seen as uncertainty measures over other uncertainty measures.

## 3.1 Probability Rankings

We now show that g-coherence and g-coherent entailment can be defined in terms of admissible probability rankings. We first give some preparative definitions.

In the sequel, we use $\alpha > 0$ to abbreviate the probabilistic formula $\neg(\alpha|\top)[0,0]$. A *probability ranking* $\kappa$ maps each probabilistic interpretation on $\mathcal{I}_\Phi$ to a member of $\{0, 1, \ldots\} \cup \{\infty\}$ such that $\kappa(Pr) = 0$ for at least one interpretation $Pr$. It is extended to all logical constraints and probabilistic formulas $F$ as follows. If $F$ is satisfiable, then $\kappa(F) = \min \{\kappa(Pr) \mid Pr \models F\}$; otherwise, $\kappa(F) = \infty$. A probability ranking $\kappa$ is *admissible* with a probabilistic knowledge base $KB = (L, P)$ iff $\kappa(\neg F) = \infty$ for all $F \in L$ and $\kappa(\phi > 0) < \infty$ and $\kappa(\phi > 0 \land (\psi|\phi)[l,u]) < \kappa(\phi > 0 \land \neg(\psi|\phi)[l,u])$ for all $(\psi|\phi)[l,u] \in P$.

The following result shows that g-coherence is equivalent to the existence of an admissible probability ranking.

**Theorem 3.1** *Let $KB = (L, P)$ be a probabilistic knowledge base. Then, $KB$ is g-coherent iff there exists a probability ranking $\kappa$ that is admissible with $KB$.*

The next theorem shows that g-coherent entailment can be expressed in terms of admissible probability rankings.

**Theorem 3.2** *Let $KB = (L, P)$ be a g-coherent probabilistic knowledge base, and let $(\psi|\phi)[l,u]$ be a conditional constraint. Then, $KB \|\!\sim^g (\psi|\phi)[l,u]$ iff $\kappa(\phi > 0) = \infty$ or $\kappa(\phi > 0 \land (\psi|\phi)[l,u]) < \kappa(\phi > 0 \land \neg(\psi|\phi)[l,u])$ for every probability ranking $\kappa$ that is admissible with $KB$.*

## 3.2 System Z

We now extend Pearl's entailment in System $Z$ [48, 32] to g-coherent probabilistic knowledge bases $KB = (L, P)$. The new notion of entailment in System $Z$ is linked to an ordered partition of $P$, a conditional constraint ranking $z$ on $KB$, and a probability ranking $\kappa^z$.

The *z-partition* of $KB$ is the unique ordered partition $(P_0, \ldots, P_k)$ of $P$ such that each $P_i$, $i \in \{0, \ldots, k\}$, is the set of all $C \in P - (P_0 \cup \cdots \cup P_{i-1})$ that is tolerated under $L$ by $P - (P_0 \cup \cdots \cup P_{i-1})$.

**Example 3.3** The $z$-partition of the probabilistic knowledge base $KB = (L, P)$ in Example 2.1 is given by:

$$(P_0) = (\{(\mathsf{have\_legs} \mid \mathsf{bird})[1,1], (\mathsf{fly} \mid \mathsf{bird})[1,1]\}).$$

The $z$-partition of the probabilistic knowledge base $KB' = (L', P')$ in Example 2.1 is given as follows:

$$(P_0', P_1') = (\{(\mathsf{have\_legs} \mid \mathsf{bird})[1,1], (\mathsf{fly} \mid \mathsf{bird})[1,1]\},$$
$$\{(\mathsf{fly} \mid \mathsf{penguin})[0, 0.05]\}). \quad \Box$$

We next define the conditional constraint ranking $z$ and the probability ranking $\kappa^z$. For every $j \in \{0, \ldots, k\}$, each $C \in P_j$ is assigned the value $j$ under $z$. The probability ranking $\kappa^z$ on all probabilistic interpretations $Pr$ is then defined as follows:

$$\kappa^z(Pr) = \begin{cases} \infty & \text{if } Pr \not\models L \\ 0 & \text{if } Pr \models L \cup P \\ 1 + \max_{C \in P: \, Pr \not\models C} z(C) & \text{otherwise.} \end{cases}$$

The following lemma shows that, in fact, $z$ is a conditional constraint ranking on $KB$ that is admissible with $KB$, and $\kappa^z$ is a probability ranking that is admissible with $KB$.

**Lemma 3.4** *Let $KB$ be a g-coherent probabilistic knowledge base. Then,*
*(a) $z$ is admissible with $KB$.*
*(b) $\kappa^z$ is admissible with $KB$.*

We next define a preference relation on probabilistic interpretations as follows. For probabilistic interpretations $Pr$ and $Pr'$, we say $Pr$ is *z-preferable* to $Pr'$ iff $\kappa^z(Pr) < \kappa^z(Pr')$. A model $Pr$ of a set of logical and conditional constraints $\mathcal{F}$ is a *z-minimal model* of $\mathcal{F}$ iff no model of $\mathcal{F}$ is $z$-preferable to $Pr$.

We finally define the notion of *z-entailment* as follows. A conditional constraint $C$ is a *z-consequence* of $KB$, denoted $KB \|\!\sim^z C$, iff every $z$-minimal model of $L$ satisfies $C$. A conditional constraint $(\psi|\phi)[l,u]$ is a *tight z-consequence* of $KB$, denoted $KB \|\!\sim^z_{tight} (\psi|\phi)[l,u]$, iff $l$ (resp., $u$) is the infimum (resp., supremum) of $Pr(\psi|\phi)$ subject to all $z$-minimal models $Pr$ of $L$ with $Pr(\phi) > 0$.

The following example shows that $z$-entailment realizes an inheritance of logical properties from classes to non-exceptional subclasses. However, logical properties are not inherited from classes to subclasses that are exceptional with respect to some other property (in classical default reasoning from conditional knowledge bases, we call this undesirable feature *inheritance blocking*).

**Example 3.5** The probabilistic knowledge base $KB$ of Example 2.1 has the following tight $z$-consequences:

$$(\mathsf{fly}|\mathsf{bird})[1,1], (\mathsf{have\_legs}|\mathsf{bird})[1,1],$$
$$(\mathsf{fly}|\mathsf{penguin})[1,1], (\mathsf{have\_legs}|\mathsf{penguin})[1,1].$$

The probabilistic knowledge base $KB'$ of Example 2.1 has the following tight $z$-consequences:

$$(\mathsf{fly}|\mathsf{bird})[1,1], (\mathsf{have\_legs}|\mathsf{bird})[1,1],$$
$$(\mathsf{fly}|\mathsf{penguin})[0, 0.05], (\mathsf{have\_legs}|\mathsf{penguin})[0,1].$$

Here, the last interval "[0, 1]" is due to the fact that the logical property of having legs is not inherited from birds to its exceptional subclass penguins. □

The following lemma describes an alternative way of defining the notion of $z$-consequence.

**Lemma 3.6** *Let $KB$ be a g-coherent probabilistic knowledge base, and let $C = (\psi|\phi)[l, u]$ be a conditional constraint. Then, $KB \hspace{0.1em}\|\hspace{-0.1em}\sim^z C$ iff $\kappa^z(\phi > 0) = \infty$ or $\kappa^z(\phi > 0 \wedge C) < \kappa^z(\phi > 0 \wedge \neg C)$.*

### 3.3 Lexicographic Entailment

We next extend Lehmann's lexicographic entailment [37] to g-coherent probabilistic knowledge bases $KB = (L, P)$. Note that even though we do not use probability rankings here, the new notion of entailment can be easily expressed by a unique single probability ranking.

We use the $z$-partition $(P_0, \ldots, P_k)$ of $KB$ to define a lexicographic preference relation on probabilistic interpretations as follows. For probabilistic interpretations $Pr$ and $Pr'$, we say $Pr$ is *lexicographically preferable* (or *lex-preferable*) to $Pr'$ iff some $i \in \{0, \ldots, k\}$ exists such that $|\{C \in P_i \mid Pr \models C\}| > |\{C \in P_i \mid Pr' \models C\}|$ and $|\{C \in P_j \mid Pr \models C\}| = |\{C \in P_j \mid Pr' \models C\}|$ for all $i < j \leq k$. A model $Pr$ of a set of logical and conditional constraints $\mathcal{F}$ is a *lexicographically minimal model* (or *lex-minimal model*) of $\mathcal{F}$ iff no model of $\mathcal{F}$ is *lex*-preferable to $Pr$.

A conditional constraint $C$ is a *lex-consequence* of $KB$, denoted $KB \hspace{0.1em}\|\hspace{-0.1em}\sim^{lex} C$, iff every *lex*-minimal model of $L$ satisfies $C$. A conditional constraint $(\psi|\phi)[l, u]$ is a *tight lex-consequence* of $KB$, denoted $KB \hspace{0.1em}\|\hspace{-0.1em}\sim^{lex}_{tight} (\psi|\phi)[l, u]$, iff $l$ (resp., $u$) is the infimum (resp., supremum) of $Pr(\psi|\phi)$ subject to all *lex*-minimal models $Pr$ of $L$ with $Pr(\phi) > 0$.

The following example shows that *lex*-entailment realizes a correct inheritance of logical properties. In particular, it does not show the problem of inheritance blocking.

**Example 3.7** The probabilistic knowledge base $KB$ of Example 2.1 has the following tight *lex*-consequences:

(fly | bird)$[1, 1]$, (have_legs | bird)$[1, 1]$,
(fly | penguin)$[1, 1]$, (have_legs | penguin)$[1, 1]$.

The probabilistic knowledge base $KB'$ of Example 2.1 has the following tight *lex*-consequences:

(fly | bird)$[1, 1]$, (have_legs | bird)$[1, 1]$,
(fly | penguin)$[0, 0.05]$, (have_legs | penguin)$[1, 1]$. □

## 4 Properties

In this section, we analyze some properties of logical entailment, g-coherent entailment, $z$-entailment, and *lex*-entailment. We first describe some general nonmonotonic properties. We then explore the relationship between the formalisms, and the one to their classical counterparts.

### 4.1 General Nonmonotonic Properties

We now analyze some general nonmonotonic properties of the probabilistic formalisms of this paper. It is important to point out that these properties are especially related to the inheritance of logical knowledge.

We first consider the postulates *Right Weakening (RW)*, *Reflexivity (Ref)*, *Left Logical Equivalence (LLE)*, *Cut*, *Cautious Monotonicity (CM)*, and *Or* by Kraus et al. [34], which are commonly regarded as being particularly desirable for any reasonable notion of nonmonotonic entailment. The following result shows that logical entailment, g-coherent entailment, $z$-entailment, and *lex*-entailment all satisfy (probabilistic versions of) these postulates.

**Theorem 4.1** $\models$, $\hspace{0.1em}\|\hspace{-0.1em}\sim^g$, $\hspace{0.1em}\|\hspace{-0.1em}\sim^z$, and $\hspace{0.1em}\|\hspace{-0.1em}\sim^{lex}$ *satisfy the following properties for all probabilistic knowledge bases $KB = (L, P)$, all events $\varepsilon, \varepsilon', \phi$, and $\psi$, and all $l, l', u, u' \in [0, 1]$:*

*RW.* If $(\phi|\top)[l, u] \Rightarrow (\psi|\top)[l', u']$ is logically valid and $KB \hspace{0.1em}\|\hspace{-0.1em}\sim (\phi|\varepsilon)[l, u]$, then $KB \hspace{0.1em}\|\hspace{-0.1em}\sim (\psi|\varepsilon)[l', u']$.

*Ref.* $KB \hspace{0.1em}\|\hspace{-0.1em}\sim (\varepsilon|\varepsilon)[1, 1]$.

*LLE.* If $\varepsilon \Leftrightarrow \varepsilon'$ is logically valid, then $KB \hspace{0.1em}\|\hspace{-0.1em}\sim (\phi|\varepsilon)[l, u]$ iff $KB \hspace{0.1em}\|\hspace{-0.1em}\sim (\phi|\varepsilon')[l, u]$.

*Cut.* If $KB \hspace{0.1em}\|\hspace{-0.1em}\sim (\varepsilon|\varepsilon')[1, 1]$ and $KB \hspace{0.1em}\|\hspace{-0.1em}\sim (\phi|\varepsilon \wedge \varepsilon')[l, u]$, then $KB \hspace{0.1em}\|\hspace{-0.1em}\sim (\phi|\varepsilon')[l, u]$.

*CM.* If $KB \hspace{0.1em}\|\hspace{-0.1em}\sim (\varepsilon|\varepsilon')[1, 1]$ and $KB \hspace{0.1em}\|\hspace{-0.1em}\sim (\phi|\varepsilon')[l, u]$, then $KB \hspace{0.1em}\|\hspace{-0.1em}\sim (\phi|\varepsilon \wedge \varepsilon')[l, u]$.

*Or.* If $KB \hspace{0.1em}\|\hspace{-0.1em}\sim (\phi|\varepsilon)[1, 1]$ and $KB \hspace{0.1em}\|\hspace{-0.1em}\sim (\phi|\varepsilon')[1, 1]$, then $KB \hspace{0.1em}\|\hspace{-0.1em}\sim (\phi|\varepsilon \vee \varepsilon')[1, 1]$.

Another desirable property is *Rational Monotonicity (RM)* [34], which describes a restricted form of monotony, and allows to ignore certain kinds of irrelevant knowledge. The next theorem shows that logical entailment, $z$-entailment, and *lex*-entailment satisfy *RM*. Here, we use $KB \hspace{0.1em}\|\hspace{-0.1em}\not\sim C$ to denote that it is not the case that $KB \hspace{0.1em}\|\hspace{-0.1em}\sim C$.

**Theorem 4.2** $\models$, $\hspace{0.1em}\|\hspace{-0.1em}\sim^z$, and $\hspace{0.1em}\|\hspace{-0.1em}\sim^{lex}$ *satisfy the following property for all probabilistic knowledge bases $KB = (L, P)$, all events $\varepsilon, \varepsilon'$, and $\psi$, and all $l, u \in [0, 1]$:*

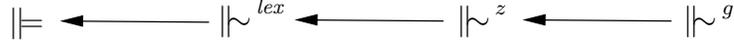

Figure 1: Logical Implications between Consequence Relations

*RM.* If $KB \;\|\!\!\sim (\psi|\varepsilon)[1,1]$ and $KB \;\|\!\!\not\sim (\neg\varepsilon'|\varepsilon)[1,1]$, then $KB \;\|\!\!\sim (\psi|\varepsilon \wedge \varepsilon')[1,1]$.

The following example shows that g-coherent entailment, however, generally does not satisfy *RM*.

**Example 4.3** Consider the probabilistic knowledge base

$$KB = (L, P) = (\{\text{bird} \Leftarrow \text{eagle}\}, \{(\text{fly} \,|\, \text{bird})[1,1]\})\,.$$

Then, $(\text{fly} \,|\, \text{bird})[1,1]$ is a logical consequence (resp., g-coherent consequence, $z$-consequence, $lex$-consequence) of $KB$. Moreover, $(\neg\text{eagle} \,|\, \text{bird})[1,1]$ is not a logical consequence (resp., g-coherent consequence, $z$-consequence, $lex$-consequence) of $KB$.

Observe now that $(\text{fly} \,|\, \text{bird} \wedge \text{eagle})[1,1]$ is a logical consequence (resp., $z$-consequence, $lex$-consequence) of $KB$, but $(\text{fly} \,|\, \text{bird} \wedge \text{eagle})[1,1]$ is not a g-coherent consequence of $KB$. We remark that $(\text{fly} \,|\, \text{bird} \wedge \text{eagle})[1,1]$ is the tight logical consequence (resp., $z$-consequence, $lex$-consequence) of $KB$, while $(\text{fly} \,|\, \text{bird} \wedge \text{eagle})[0,1]$ is the tight g-coherent consequence of $KB$. □

We now consider the property *Irrelevance (Irr)* adapted from [7]. Informally, *Irr* says that $\varepsilon'$ is irrelevant to a conclusion "$P \;\|\!\!\sim (\psi|\varepsilon)[1,1]$" when they are defined over disjoint sets of atoms. The following result shows that logical entailment, $z$-entailment, and $lex$-entailment satisfy *Irr*.

**Theorem 4.4** $\models$, $\|\!\!\sim^z$, and $\|\!\!\sim^{lex}$ satisfy the following property for all probabilistic knowledge bases $KB = (L, P)$ and all events $\varepsilon$, $\varepsilon'$, and $\psi$:

*Irr.* If $KB \;\|\!\!\sim (\psi|\varepsilon)[1,1]$, and no atom of $KB$ and $(\psi|\varepsilon)[1,1]$ occurs in $\varepsilon'$, then $KB \;\|\!\!\sim (\psi|\varepsilon \wedge \varepsilon')[1,1]$.

The following example shows that g-coherent entailment, however, does not satisfy *Irr*.

**Example 4.5** Consider the probabilistic knowledge base

$$KB = (L, P) = (\emptyset, \{(\text{fly} \,|\, \text{bird})[1,1]\})\,.$$

Clearly, $(\text{fly} \,|\, \text{bird})[1,1]$ is a logical consequence (resp., g-coherent consequence, $z$-consequence, $lex$-consequence) of $KB$. Observe now that $(\text{fly} \,|\, \text{red} \wedge \text{bird})[1,1]$ is a logical consequence (resp., $z$-consequence, $lex$-consequence) of $KB$, but $(\text{fly} \,|\, \text{red} \wedge \text{bird})[1,1]$ is not a g-coherent consequence of $KB$. We remark that $(\text{fly} \,|\, \text{red} \wedge \text{bird})[1,1]$ is the tight logical consequence (resp., $z$-consequence, $lex$-consequence) of $KB$, while $(\text{fly} \,|\, \text{red} \wedge \text{bird})[0,1]$ is the tight g-coherent consequence of $KB$. □

We finally consider the property *Direct Inference (DI)* adapted from [3]. Informally, *DI* expresses that $P$ should entail all its own conditional constraints. The following theorem shows that logical entailment, g-coherent entailment, $z$-entailment, and $lex$-entailment all satisfy *DI*.

**Theorem 4.6** $\models$, $\|\!\!\sim^g$, $\|\!\!\sim^z$, and $\|\!\!\sim^{lex}$ satisfy the following property for all probabilistic knowledge bases $KB = (L, P)$, all events $\varepsilon$, $\phi$, and $\psi$, and all $l, u \in [0,1]$:

*DI.* If $(\psi|\phi)[l,u] \in P$ and $\varepsilon \Leftrightarrow \phi$ is logically valid, then $KB \;\|\!\!\sim (\psi|\varepsilon)[l,u]$.

### 4.2 Relationship between Probabilistic Formalisms

We now explore the logical implications between the probabilistic formalisms of this paper.

The following theorem shows that the logical implications illustrated in Fig. 1 hold between logical entailment, g-coherent entailment, $z$-entailment, and $lex$-entailment.

**Theorem 4.7** Let $KB = (L, P)$ be a g-coherent probabilistic knowledge base, and let $C = (\psi|\phi)[l, u]$ be a conditional constraint. Then,

(a) $KB \;\|\!\!\sim^g C$ implies $KB \;\|\!\!\sim^z C$.

(b) $KB \;\|\!\!\sim^z C$ implies $KB \;\|\!\!\sim^{lex} C$.

(c) $KB \;\|\!\!\sim^{lex} C$ implies $KB \models C$.

In general, none of the converse implications holds, as Examples 2.2, 2.3, 3.5, and 3.7 immediately show. But in the special case when $L \cup P$ has a model in which the conditioning event $\phi$ has a positive probability, the notions of logical entailment, $z$-entailment, and $lex$-entailment of $(\psi|\phi)[l,u]$ from $KB$ all coincide. This important result is expressed by the following theorem.

**Theorem 4.8** Let $KB = (L, P)$ be a g-coherent probabilistic knowledge base, and let $C = (\psi|\phi)[l, u]$ be a

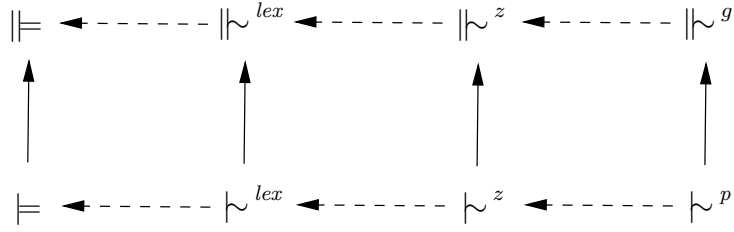

Figure 2: Relationship to Classical Formalisms

*conditional constraint such that $L \cup P$ has a model $Pr$ with $Pr(\phi) > 0$. Then,*

$$KB \models C \text{ iff } KB \mathrel{\Vert\!\sim}^{lex} C \text{ iff } KB \mathrel{\Vert\!\sim}^{z} C.$$

The following example shows that g-coherent entailment, however, generally does not coincide with logical entailment when $L \cup P$ has a model $Pr$ with $Pr(\phi) > 0$.

**Example 4.9** Consider again the probabilistic knowledge base $KB = (L, P) = (\{\text{bird} \Leftarrow \text{eagle}\}, \{(\text{fly} \mid \text{bird})[1, 1]\})$. Then, $L \cup P$ has clearly a model $Pr$ with $Pr(\text{eagle}) > 0$.

It is now easy to verify that $(\text{fly} \mid \text{eagle})[1, 1]$ is a logical consequence (resp., $z$-consequence, $lex$-consequence) of $KB$, but $(\text{fly} \mid \text{eagle})[1, 1]$ is not a g-coherent consequence of $KB$. We remark that $(\text{fly} \mid \text{eagle})[1, 1]$ is the tight logical consequence (resp., $z$-consequence, $lex$-consequence) of $KB$, while $(\text{fly} \mid \text{eagle})[0, 1]$ is the tight g-coherent consequence of $KB$. □

### 4.3 Relationship to Classical Formalisms

We now analyze the relationship between the probabilistic formalisms of this paper and their classical counterparts. We first give some technical preparation as follows.

The operator $\gamma$ on conditional constraints, sets of conditional constraints, and conditional knowledge bases replaces each conditional constraint $(\psi \mid \phi)[1, 1]$ by the classical default $\psi \leftarrow \phi$. We use the expressions $\mathrel{\mid\!\sim}^{p}$, $\mathrel{\mid\!\sim}^{z}$, $\mathrel{\mid\!\sim}^{lex}$, and $\models$ to denote the classical notions of entailment in System $P$, Pearl's entailment in System $Z$, Lehmann's lexicographic entailment, and logical entailment, respectively.

The following well-known result says that logical entailment in model-theoretic probabilistic logic is a generalization of logical entailment in classical propositional logic.

**Theorem 4.10** *Let $KB = (L, P)$ be a g-coherent probabilistic knowledge base, where $P = \{(\psi_1 \mid \phi_1)[1, 1], \ldots, (\psi_n \mid \phi_n)[1, 1]\}$, and let $(\beta \mid \alpha)[1, 1]$ be a conditional constraint. Then,*

$$KB \models (\beta \mid \alpha)[1, 1] \text{ iff } \gamma(KB) \models \beta \Leftarrow \alpha.$$

The next theorem recalls the recent result by Biazzo et al. [10] that g-coherent entailment is a generalization of default entailment in System $P$.

**Theorem 4.11 (Biazzo et al. [10])** *Let $KB = (L, P)$ be a g-coherent probabilistic knowledge base, where $P = \{(\psi_1 \mid \phi_1)[1, 1], \ldots, (\psi_n \mid \phi_n)[1, 1]\}$, and let $(\beta \mid \alpha)[1, 1]$ be a conditional constraint. Then,*

$$KB \mathrel{\Vert\!\sim}^{g} (\beta \mid \alpha)[1, 1] \text{ iff } \gamma(KB) \mathrel{\mid\!\sim}^{p} \beta \leftarrow \alpha.$$

The following theorem shows that the new notions of $z$- and $lex$-entailment for g-coherent probabilistic knowledge bases generalize their classical counterparts for $\varepsilon$-consistent conditional knowledge bases.

**Theorem 4.12** *Let $KB = (L, P)$ be a g-coherent probabilistic knowledge base, where $P = \{(\psi_1 \mid \phi_1)[1, 1], \ldots, (\psi_n \mid \phi_n)[1, 1]\}$, and let $(\beta \mid \alpha)[1, 1]$ be a conditional constraint. Then,*

*(a) $KB \mathrel{\Vert\!\sim}^{z} (\beta \mid \alpha)[1, 1]$ iff $\gamma(KB) \mathrel{\mid\!\sim}^{z} \beta \leftarrow \alpha$.*

*(b) $KB \mathrel{\Vert\!\sim}^{lex} (\beta \mid \alpha)[1, 1]$ iff $\gamma(KB) \mathrel{\mid\!\sim}^{lex} \beta \leftarrow \alpha$.*

## 5 Summary and Outlook

We have presented probabilistic generalizations of Pearl's entailment in System $Z$ and of Lehmann's lexicographic entailment that lie between entailment in model-theoretic probabilistic logic and entailment in probabilistic logic under coherence. That is, the new formalisms properly generalize their counterparts in classical default reasoning, they are weaker than model-theoretic probabilistic entailment, and they are stronger than probabilistic entailment under coherence. The new formalisms are useful especially for handling probabilistic inconsistencies related to conditioning on zero events. Furthermore, they can be applied for probabilistic belief revision. More generally, in the same spirit as a similar previous paper, this paper sheds light on exciting new formalisms for probabilistic reasoning beyond the well-known standard ones.

An interesting topic of future research is to develop algorithms for the new formalisms and to analyze their computational complexity (along the lines of [44]).

Another very interesting topic of future research is to develop and explore further nonmonotonic formalisms for reasoning with conditional constraints. Besides extending classical formalisms for default reasoning from conditional knowledge bases, which may additionally contain a strength assignment to the defaults, one may also think about combining the new formalisms of this paper and of [44] with some probability selection technique (as e.g. maximum entropy or center of mass).

**Acknowledgments**

This work has been supported by a Marie Curie Individual Fellowship of the European Community and by the Austrian Science Fund Project Z29-INF. I am grateful to Angelo Gilio and Lluis Godo for their useful comments.

**References**

[1] E. W. Adams. *The Logic of Conditionals*, volume 86 of *Synthese Library*. D. Reidel, Dordrecht, Netherlands, 1975.

[2] S. Amarger, D. Dubois, and H. Prade. Constraint propagation with imprecise conditional probabilities. In *Proceedings UAI-91*, pages 26–34. Morgan Kaufmann, 1991.

[3] F. Bacchus, A. Grove, J. Halpern, and D. Koller. From statistical knowledge bases to degrees of belief. *Artif. Intell.*, 87:75–143, 1996.

[4] S. Benferhat, C. Cayrol, D. Dubois, J. Lang, and H. Prade. Inconsistency management and prioritized syntax-based entailment. In *Proceedings IJCAI-93*, pages 640–645. Morgan Kaufmann, 1993.

[5] S. Benferhat, D. Dubois, and H. Prade. Representing default rules in possibilistic logic. In *Proceedings KR-92*, pages 673–684. Morgan Kaufmann, 1992.

[6] S. Benferhat, D. Dubois, and H. Prade. Nonmonotonic reasoning, conditional objects and possibility theory. *Artif. Intell.*, 92(1–2):259–276, 1997.

[7] S. Benferhat, A. Saffiotti, and P. Smets. Belief functions and default reasoning. *Artif. Intell.*, 122(1–2):1–69, 2000.

[8] V. Biazzo and A. Gilio. A generalization of the fundamental theorem of de Finetti for imprecise conditional probability assessments. *Int. J. Approx. Reasoning*, 24(2–3):251–272, 2000.

[9] V. Biazzo, A. Gilio, T. Lukasiewicz, and G. Sanfilippo. Probabilistic logic under coherence: Complexity and algorithms. In *Proceedings ISIPTA-01*, pages 51–61, 2001.

[10] V. Biazzo, A. Gilio, T. Lukasiewicz, and G. Sanfilippo. Probabilistic logic under coherence, model-theoretic probabilistic logic, and default reasoning. In *Proceedings ECSQARU-01*, volume 2143 of *LNCS/LNAI*, pages 290–302. Springer, 2001.

[11] G. Boole. *An Investigation of the Laws of Thought, on which are Founded the Mathematical Theories of Logic and Probabilities*. Walton and Maberley, London, 1854. (reprint: Dover Publications, New York, 1958).

[12] G. Coletti. Coherent numerical and ordinal probabilistic assessments. *IEEE Trans. Syst. Man Cybern.*, 24(12):1747–1754, 1994.

[13] G. Coletti and R. Scozzafava. Characterization of coherent conditional probabilities as a tool for their assessment and extension. *Journal of Uncertainty, Fuzziness and Knowledge-based Systems*, 4(2):103–127, 1996.

[14] G. Coletti and R. Scozzafava. Coherent upper and lower Bayesian updating. In *Proceedings ISIPTA-99*, pages 101–110, 1999.

[15] G. Coletti and R. Scozzafava. Conditioning and inference in intelligent systems. *Soft Computing*, 3(3):118–130, 1999.

[16] D. Dubois and H. Prade. On fuzzy syllogisms. *Computational Intelligence*, 4(2):171–179, 1988.

[17] D. Dubois and H. Prade. Possibilistic logic, preferential models, non-monotonicity and related issues. In *Proceedings IJCAI-91*, pages 419–424. Morgan Kaufmann, 1991.

[18] D. Dubois and H. Prade. Conditional objects as nonmonotonic consequence relationships. *IEEE Trans. Syst. Man Cybern.*, 24(12):1724–1740, 1994.

[19] D. Dubois, H. Prade, L. Godo, and R. L. de Màntaras. Qualitative reasoning with imprecise probabilities. *Journal of Intelligent Information Systems*, 2:319–363, 1993.

[20] D. Dubois, H. Prade, and J.-M. Touscas. Inference with imprecise numerical quantifiers. In Z. W. Ras and M. Zemankova, editors, *Intelligent Systems*, chapter 3, pages 53–72. Ellis Horwood, 1990.

[21] T. Eiter and T. Lukasiewicz. Default reasoning from conditional knowledge bases: Complexity and tractable cases. *Artif. Intell.*, 124(2):169–241, 2000.

[22] R. Fagin, J. Y. Halpern, and N. Megiddo. A logic for reasoning about probabilities. *Inf. Comput.*, 87:78–128, 1990.

[23] A. M. Frisch and P. Haddawy. Anytime deduction for probabilistic logic. *Artif. Intell.*, 69:93–122, 1994.

[24] D. M. Gabbay and P. Smets, editors. *Handbook on Defeasible Reasoning and Uncertainty Management Systems*. Kluwer Academic, Dordrecht, Netherlands, 1998.

[25] H. Geffner. *Default Reasoning: Causal and Conditional Theories*. MIT Press, 1992.

[26] H. Geffner and J. Pearl. Conditional entailment: Bridging two approaches to default reasoning. *Artif. Intell.*, 53(2–3):209–244, 1992.

[27] A. Gilio. Probabilistic consistency of conditional probability bounds. In *Advances in Intelligent Computing*, volume 945 of *LNCS*, pages 200–209. Springer, 1995.

[28] A. Gilio. Probabilistic reasoning under coherence in System P. *Ann. Math. Artif. Intell.*, 34(1–3):5–34, 2002.

[29] A. Gilio and R. Scozzafava. Conditional events in probability assessment and revision. *IEEE Trans. Syst. Man Cybern.*, 24(12):1741–1746, 1994.

[30] M. Goldszmidt, P. Morris, and J. Pearl. A maximum entropy approach to nonmonotonic reasoning. *IEEE Trans. Pattern Anal. Mach. Intell.*, 15(3):220–232, 1993.

[31] M. Goldszmidt and J. Pearl. Rank-based systems: A simple approach to belief revision, belief update and reasoning about evidence and actions. In *Proceedings KR-92*, pages 661–672. Morgan Kaufmann, 1992.

[32] M. Goldszmidt and J. Pearl. Qualitative probabilities for default reasoning, belief revision, and causal modeling. *Artif. Intell.*, 84(1–2):57–112, 1996.

[33] J. Heinsohn. Probabilistic description logics. In *Proceedings UAI-94*, pages 311–318. Morgan Kaufmann, 1994.

[34] S. Kraus, D. Lehmann, and M. Magidor. Nonmonotonic reasoning, preferential models and cumulative logics. *Artif. Intell.*, 14(1):167–207, 1990.

[35] P. Lamarre. A promenade from monotonicity to non-monotonicity following a theorem prover. In *Proceedings KR-92*, pages 572–580. Morgan Kaufmann, 1992.

[36] D. Lehmann. What does a conditional knowledge base entail? In *Proceedings KR-89*, pages 212–222. Morgan Kaufmann, 1989.

[37] D. Lehmann. Another perspective on default reasoning. *Ann. Math. Artif. Intell.*, 15(1):61–82, 1995.

[38] D. Lehmann and M. Magidor. What does a conditional knowledge base entail? *Artif. Intell.*, 55(1):1–60, 1992.

[39] T. Lukasiewicz. Local probabilistic deduction from taxonomic and probabilistic knowledge-bases over conjunctive events. *Int. J. Approx. Reasoning*, 21(1):23–61, 1999.

[40] T. Lukasiewicz. Probabilistic deduction with conditional constraints over basic events. *J. Artif. Intell. Res.*, 10:199–241, 1999.

[41] T. Lukasiewicz. Probabilistic logic programming under inheritance with overriding. In *Proceedings UAI-01*, pages 329–336. Morgan Kaufmann, 2001.

[42] T. Lukasiewicz. Probabilistic logic programming with conditional constraints. *ACM Trans. Computat. Logic*, 2(3):289–339, 2001.

[43] T. Lukasiewicz. Nonmonotonic probabilistic logics between model-theoretic probabilistic logic and probabilistic logic under coherence. Technical Report INFSYS RR-1843-02-02, Institut für Informationssysteme, TU Wien, 2002.

[44] T. Lukasiewicz. Probabilistic default reasoning with conditional constraints. *Ann. Math. Artif. Intell.*, 34(1–3):35–88, 2002.

[45] C. Luo, C. Yu, J. Lobo, G. Wang, and T. Pham. Computation of best bounds of probabilities from uncertain data. *Computational Intelligence*, 12(4):541–566, 1996.

[46] N. J. Nilsson. Probabilistic logic. *Artif. Intell.*, 28:71–88, 1986.

[47] J. Pearl. Probabilistic semantics for nonmonotonic reasoning: A survey. In *Proceedings KR-89*, pages 505–516. Morgan Kaufmann, 1989.

[48] J. Pearl. System Z: A natural ordering of defaults with tractable applications to default reasoning. In *Proceedings TARK-90*, pages 121–135. Morgan Kaufmann, 1990.

[49] R. Pelessoni and P. Vicig. A consistency problem for imprecise conditional probability assessments. In *Proceedings IPMU-98*, pages 1478–1485, 1998.

[50] R. Scozzafava. Subjective conditional probability and coherence principles for handling partial information. *Mathware Soft Comput.*, 3(1):183–192, 1996.

[51] Y. Shoham. A semantical approach to nonmonotonic logics. In *Proceedings of the 2nd IEEE Symposium on Logic in Computer Science*, pages 275–279, 1987.

[52] W. Spohn. Ordinal conditional functions: A dynamic theory of epistemic states. In W. Harper and B. Skyrms, editors, *Causation in Decision, Belief Change, and Statistics*, volume 2, pages 105–134. Reidel, Dordrecht, Netherlands, 1988.

[53] H. Thöne, U. Güntzer, and W. Kießling. Towards precision of probabilistic bounds propagation. In *Proceedings UAI-92*, pages 315–322. Morgan Kaufmann, 1992.

[54] P. Walley. *Statistical Reasoning with Imprecise Probabilities*. Chapman and Hall, 1991.